\documentclass[a4paper,twoside]{article}

\usepackage{epsfig}
\usepackage{subcaption}
\usepackage{calc}
\usepackage{amssymb}
\usepackage{amstext}
\usepackage{amsmath}
\usepackage{amsthm}
\usepackage{multicol}
\usepackage{pslatex}
\usepackage{apalike}
\usepackage[bottom]{footmisc}
\usepackage[hidelinks]{hyperref}
\usepackage{xcolor}
\usepackage{fancyvrb}
\usepackage{listings}
\usepackage[relative,overlay]{textpos}
\usepackage{tikz}
\usepackage{graphicx}
\usepackage{SCITEPRESS}     

\newcommand*\circledorange[1]{\tikz[baseline=(char.base)]{
    \node[shape=circle,fill=orange,text=white,inner sep=2pt] (char) {#1};}}

\lstdefinestyle{base}{
  language=Python,
  emptylines=1,
  breaklines=true,
  basicstyle=\ttfamily\color{black},
  moredelim=**[is][\color{red}]{@}{@},
  moredelim=**[is][\color{blue}]{|}{|}
}

\begin{document}

\title{Large Language Models (GPT) Struggle to Answer Multiple-Choice Questions about Code}

\author{\authorname{Jaromir Savelka\sup{1}\orcidAuthor{0000-0002-3674-5456}, Arav Agarwal\sup{1}\orcidAuthor{0000-0001-9848-1663}, Christopher Bogart\sup{1}\orcidAuthor{0000-0001-8581-115X} and Majd Sakr\sup{1}\orcidAuthor{0000-0001-5150-8259}}
\affiliation{\sup{1}School of Computer Science, Carnegie Mellon University, Pittsburgh, PA, USA}
}


\abstract{We analyzed effectiveness of three generative pre-trained transformer (GPT) models in answering multiple-choice question~(MCQ) assessments, often involving short snippets of code, from introductory and intermediate programming courses at the postsecondary level. 
This emerging technology stirs countless discussions of its potential uses (e.g., exercise generation, code explanation) as well as misuses in programming education~(e.g., cheating). However, the capabilities of GPT models and their limitations to reason about and/or analyze code in educational settings have been under-explored. We evaluated several OpenAI's GPT models on formative and summative MCQ assessments from three Python courses (530 questions). We found that MCQs containing code snippets are not answered as successfully as those that only contain natural language. While questions requiring to fill-in a blank in the code or completing a natural language statement about the snippet are handled rather successfully, MCQs that require analysis and/or reasoning about the code (e.g., what is true/false about the snippet, or what is its output) appear to be the most challenging. These findings can be leveraged by educators to adapt their instructional practices and assessments in programming courses, so that GPT becomes a valuable assistant for a learner as opposed to a source of confusion and/or potential hindrance in the learning process.}

\onecolumn \maketitle \normalsize \setcounter{footnote}{0} \vfill

\section{\uppercase{Introduction}}
\label{sec:introduction}
This paper analyzes the effectiveness of generative pre-trained transformers (GPT), specifically \verb|text-davinci-*| models, to handle multiple-choice question (MCQ) assessments, often involving small snippets of code, from introductory and intermediate programming courses. We manually collected a sizeable data set of 530 MCQs from three existing Python courses. Using a combination of simple pattern matching and manual curation, we organized the questions into meaningful categories according to their type (e.g., true/false questions, or questions asking about an output of the provided code snippet). We analyzed the performance of the GPT models across the categories to determine if questions of a certain type are handled more successfully than questions of other types. We also benchmark the older InstructGPT \verb|text-davinci-001| model against the more recent GPT\nobreakdash-3.5 \verb|text-davinci-002| and \verb|text-davinci-003| models to gauge the rate of improvement that has been achieved over the past several years.

There has been a burst of public attention to GPT models' potential impact on education as the result of the recent release of OpenAI's ChatGPT\footnote{ChatGPT. \url{https://chat.openai.com/} [Accessed 2023-01-26]}.  For example, the tool has been blocked by New York City public schools~\cite{ElsenRooney2023} because it may enable student plagiarism and provide inappropriate or incorrect content. Universities have also been reacting, adjusting assignments \cite{Huang2023} and seeking out tools like GPTZero that detect text generated by AI tools~\cite{Bowman2023}. OpenAI have released a similar tool themselves. However, reliability of these tools has not been thoroughly tested.

Programming instructors as well as CS educators in general have been sensitized to this development even earlier. Large language models, such as GPT, can generate computer program code (i.e., perform computer program synthesis) with a high degree of success. They can also explain computer program code in natural language terms. Recently, a number of computer program code generation tools have been released. Among these, the most prominent ones are OpenAI's Codex~\cite{https://doi.org/10.48550/arxiv.2107.03374}, DeepMind's AlphaCode~\cite{doi:10.1126/science.abq1158}, and Amazon's CodeWhisperer~\cite{ankur2022}. GitHub's Copilot\footnote{GitHub Copilot: Your AI pair programmer. Available at: \url{https://github.com/features/copilot} [Accessed 2023-01-20]} (a version of Codex) conveniently integrates with IDEs, such as Visual Studio Code, and hence has attracted much attention. Microsoft dubs Copilot as ``Your AI pair programmer'' (a reference to pair programming \cite{beck2000extreme,mcdowell2002effects}). Since it is available for free to students and educators, it is inevitable that learners will use it to complete their course assignments and assessments. Similarly, there are no technical or cost barriers to using ChatGPT which can be, among many other things, leveraged to generate answers to MCQ questions.

To investigate how GPT models handle the MCQ assessments of various types in a programming education context, we analyzed the following research questions:

\begin{itemize}
    \item Is there a difference between how successfully the GPT models handle questions that contain only natural language and those that also contain snippets of computer code?
    \item Are there particular types of MCQs that are more challenging for the GPT models compared to other types of MCQs?
\end{itemize}

By carrying out this work, we provide the following contributions to the CS education research community. To the best of our knowledge, this is the first comprehensive study that:

\begin{itemize}
    \item Evaluates the performance of GPT models on MCQ-style assessments that involve code snippets, across different types of such questions.
    \item Lays a systematic foundation for discussions about suitable uses of GPT models in programming classes by providing quantitative analysis of the models capabilities and limitations in handling of computer code.
\end{itemize}

\begin{figure*}[t!]
\includegraphics[width=\textwidth]{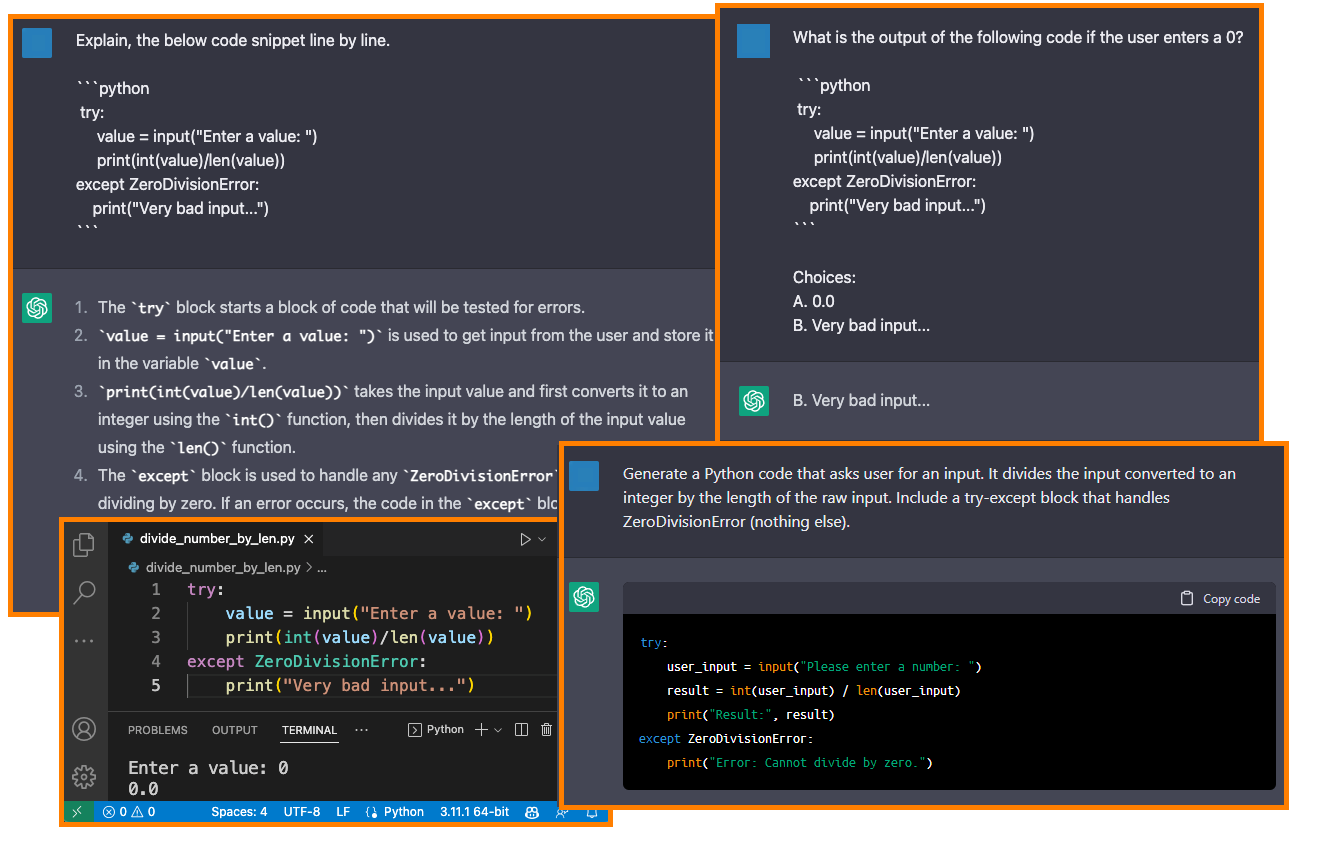}
\begin{textblock*}{3.4cm}(.5cm,-5.3cm)
\circledorange{1}
\end{textblock*}
\begin{textblock*}{3.4cm}(11.75cm,-2.15cm)
\circledorange{2}
\end{textblock*}
\begin{textblock*}{3.4cm}(11.25cm,-5.6cm)
\circledorange{3}
\end{textblock*}
\begin{textblock*}{3.4cm}(3.55cm,-1.15cm)
\circledorange{4}
\end{textblock*}
\caption{The upper-left screenshot depicts a conversation with ChatGPT when asked to explain a code snippet line by line. It correctly explains the bahavior (1). The lower-right shows a conversation with ChatGPT when asked to generate the code snippet with the same behavior. The generated code is correct (2). The upper-right screenshot depicts a conversation with ChatGPT when asked a straightforward MCQ about a code it can correctly explain line by line as well as correctly generate. The answer is wrong (3)---compare the actual output of the code snippet which is shown in the lower-left corner (4).}
\label{fig:motivation}
\end{figure*}

\section{Motivating Example}
\label{sec:motivation}
Consider the below Python script that asks a user to input a value which is expected to be a number. The entered value of type \verb|str| is cast to an \verb|int| and divided by the length of the raw input (\verb|str|). Note that the code defends against the possibility of a \verb|ZeroDivisionError| which cannot really occur, as explained below. However, this likely confuses GPT models when answering questions about this snippet.

\begin{Verbatim}
try:
    value = input("Enter a value: ")
    print(int(value) / len(value))
except ZeroDivisionError:
    print("Very bad input...")
\end{Verbatim}

\noindent If a user enters 22, then the output of the script would be 11.0 (i.e., 22 / 2). As shown in Figure \ref{fig:motivation}, if one provides ChatGPT (one of the state-of-the-art GPT-3.5 models) with the code snippet and asks, ``what would be the output if the user enters 0,'' (letting ChatGPT choose from ``A. 0.0'' or ``B. Very bad input...''), the provided answer is ``B. Very bad input...'' Of course, this is an incorrect answer because the length of the string \verb|"0"| is 1 and, hence, the output is 0.0 (as shown in Figure \ref{fig:motivation}).

A human learner making this error would likely be suspected of having several crucial misconceptions. Firstly, selecting the ``B.~Very bad input...'' option would be somewhat more understandable if the \verb|value| variable were not placed within the \verb|len()| function call. In that case, one could assume that the learner simply failed to recognize that the output of the \verb|input()| function call is a \verb|str| and assumed it was an \verb|int| instead. However, applying the \verb|len()| function to an \verb|int| would result in a \verb|TypeError| being raised. Hence, the only input that could theoretically raise a \verb|ZeroDivisionError| would be an empty string. However, even that input would not result in that particular error because it would fail on an attempt to cast the \verb|value| variable to \verb|int| (\verb|ValueError|) that would occur prior to the division. Overall, a human learner selecting the ``B. Very bad input...'' answer over the correct ``A. 0.0'' would clearly demonstrate a lack of understanding of the workings of this simple snippet of code. 

Figure \ref{fig:motivation} shows the output of ChatGPT when asked to explain the code snippet line by line. Interestingly, the explanation is correct, including the line where the division takes place. With respect to the statement on that line, it declares that: ``[it] takes the input value and first converts it to an integer using the \verb|int()| function, then divides it by the length of the input value using the \verb|len()| function.'' Furthermore, Figure \ref{fig:motivation} also shows the output of ChatGPT when asked to generate Python code with the same functionality as the provided code snippet. From the natural language description, ChatGPT generates correct Python code with the specified behavior.

In this example, a GPT model is capable of correctly explaining the behavior (execution) of a computer program on a local level (i.e., line by line). It is equally capable of generating the computer program from a natural language description. Yet, it fails spectacularly in answering simple questions about the very same program. This is quite likely in stark contrast with a typical human learner. A learner capable of independently writing the program from the natural language description as well as correctly explaining its execution line by line, would quite likely be in a position to answer such questions with ease.

\section{\uppercase{Related Work}}
\label{sec:related_work}

In prior work, we evaluated the capability of a GPT model (\verb|text-davinci-003|), to pass a diverse set of assessment instruments, including MCQs, in the realistic context of full-fledged programming courses~\cite{Savelka2023}. We found that the current GPT models are not capable of  passing the full spectrum of assessments typically involved in a Python programming course (below 70\% on even entry-level modules); but a straightforward application of these models could enable a learner to obtain a non-trivial portion of the overall available score~(over 55\%) in introductory and intermediate courses alike. We observed that an important limitation of the  GPT models is their apparent struggle with activities that require chains of reasoning steps, and that there appeared to be a difference in success rate between MCQs that contain a code snippet and those that do not \cite{Savelka2023}. In this paper, we further explore this phenomenon, focusing on discovery of more fine-grained properties of MCQs that are challenging for the GPT models to handle.

To the best of our knowledge, there is no other study of GPT's performance on MCQs from the programming domain. There is work evaluating the performance on MCQ data sets from other domains; in many cases the tool does better than random chance; sometimes even well enough to pass a test. For example, Robinson et al. apply InstructGPT \cite{ouyang2022training} and Codex to OpenBookQA~\cite{mihaylov2018can}, StoryCloze \cite{mostafazadeh2016corpus}, and RACE-m \cite{lai2017race} data sets which focus on multi-hop reasoning, recall, and reading comprehension, reporting 77.4-89.2\% accuracy \cite{robinson2022}. In some cases, GPT can generate code when applied to programming assignments in higher education courses. Drori and Verma used Codex to write Python programs to solve 60 computational linear algebra MCQs, reporting 100\% accuracy \cite{https://doi.org/10.48550/arxiv.2111.08171}. Others have used GPT models to solve various MCQ-based exams, including the United States Medical Licensing Examination (USMLE), with accuracy around 50\% \cite{kung2022performance,Gilson2022HowWD,Lievin2022CanLL}, the Multistate Bar Examination (MBE) \cite{bommarito2022gpt}, and the American Institute of Certified Public Accountants' (AICPA) Regulation (REG) exam \cite{bommarito2023gpt}.

Although, programming-related MCQs have not been studied directly, some researchers in adjacent fields have studied reasoning about similarly formal topics. Although, GPT can often answer questions \emph{about} systems and rules, it is especially challenged by tasks that involve \emph{applying} them and reasoning about their implications in novel examples. Hendryks et al. created data set that includes a wide variety of MCQs across STEM, humanities and arts, with GPT-3 performing at levels above 50\% for subjects such as marketing and foreign policy, but below 30\% for topics like formal logic~\cite{hendrycks2022}. They found that the model performed particularly poorly in quantitative subjects. For example, in Elementary Mathematics they note that GPT can answer questions \emph{about} arithmetic order of operations (e.g. that multiplications are performed before additions), it cannot correctly answer questions that require \emph{applying} this concept. They also note that GPT performance is not necessarily correlated with how advanced the topic is for humans, doing better at College Mathematics than Elementary Mathematics. Finally, they noted that GPT does poorly on tests of legal and moral reasoning~\cite{hendrycks2022}.

Lu et al. studied GPT models' performance on a large data set consisting of 21,208 MCQs on topics in natural science, social science, and language  \cite{pan2022}. They prompted the models to produce an explanation along with its answer and reported  1-3\% improvement in accuracy (74.04\%). In this work, we do not adopt the approach and, hence, leave space for future work as it appears quite promising and definitely applicable in the context of programming MCQs.

There is a growing body of related work on GPT models' capabilities in solving programming tasks by generating code. Finnie-Ansley et al. evaluated Codex on 23 programming tasks used as summative assessments in a CS1 programming course \cite{10.1145/3511861.3511863}. Denny et al. focused on the effects of prompt engineering when applying Copilot to a set of 166 exercises from the publicly available CodeCheck repository~\cite{https://doi.org/10.48550/arxiv.2210.15157}. Outside of the educational context, there have been studies exploring GPT's capabilities on competitive and interview programming tasks. Chen et al. released the HumanEval data set where Codex achieved 28.8\% success rate on the first attempt and 72.3\% when allowed 100 attempts \cite{https://doi.org/10.48550/arxiv.2107.03374}. Li et al. report Deepmind's AlphaCode performance on Codeforces competitions,\footnote{Codeforces. Available at: \url{https://codeforces.com/contests} [Accessed 2023-01-22]} achieving a 54.3\% ranking amongst 5,000 participants \cite{doi:10.1126/science.abq1158}. Karmakar et al. reported 96\% pass rate for Codex on a data set of 115 programming problems from HackerRank\footnote{HackerRank. Available at: \url{https://www.hackerrank.com/} [Accessed 2023-01-22]} \cite{Karmakar2022CodexHH}. Nguyen and Nadi reported Copilot's effectiveness on LeetCode\footnote{LeetCode. Available at: \url{https://leetcode.com/} [Accessed 2023-01-22]} problems, achieving 42\% accuracy \cite{9796235}.

Program code does more than control computer execution; it also, some argue primarily, serves as communication among developers~\cite{Knuth1984}. Since GPT is a text prediction model trained on code in the context of human discussions about it, the model's representation of code is likely to capture code's \emph{design intent} more strongly than code's \emph{formal properties}. For example, work from multiple studies suggest that models that interpret code depend heavily on function names and input variables \cite{Mohammadkhani2022ExplainableAF,robustness}. Although, models like GPT are not trained to simulate code execution, they can in many cases generate code based on natural language description of the code's intent. Researchers have reported varying success at generating code in response to programming assignments, ranging from Codex's 100\% success generating Python computational linear algebra programs \cite{https://doi.org/10.48550/arxiv.2111.08171}, to 78.3\% on some CS1 programming problems ~\cite{10.1145/3511861.3511863}, to 79\% on the CodeCheck\footnote{CodeCheck: Python Exercises. Available at: \url{https://horstmann.com/codecheck/python-questions.html} [Accessed 2022-01-22]} repository of Python programming problems~\cite{https://doi.org/10.48550/arxiv.2210.15157}.

Researchers have identified distinct cognitive processes involved in programming. Characterizing the kinds of learning necessary to teach programming, Robins et al. claim for example that the \emph{knowledge} of how programming constructs work is cognitively different from the \emph{strategy} or plan for how to build a program; and that programming \emph{comprehension} and \emph{generation} are distinct mental processes that must be taught.  Programming skill is a blend of related cognitive processes; it is not surprising that a generative model would not mimic all these processes equally well~\cite{robins_learning_2003}.

GPT's ability to answer questions intended as educational assessments naturally raises the question of its use for cheating.
Biderman and Raff noted that GPT solutions can evade plagiarism detection by code similarity tools such as MOSS~\cite{Biderman2022FoolingMD}. 
On the other hand, Wermelinger notes that while Copilot-generated solutions can typically pass some tests, they do not pass enough to get a passing grade on a typical assignment; he concludes that Copilot can be a useful springboard towards solving CS1 problems, but outside of very common stereotyped beginners' exercises, learners' substantial contribution is still required~\cite{wermelinger2023using}. Becker et al. include a broader discussion of the opportunities and challenges posed by code generating tools~\cite{becker2022programming}.

\section{\uppercase{Data Set}}
\label{sec:data_set}
We manually collected MCQ assessment exercises from three Python programming courses. \emph{Python Essentials - Part 1 (Basics)}\footnote{OpenEDG: Python Essentials - Part 1 (Basics). Available at: \url{https://edube.org/study/pe1} [Accessed 2023-01-15]} (\textbf{PE1}) aims to guide a learner from a state of complete programming illiteracy to a level of programming knowledge which allows them to design, write, debug, and run programs encoded in the Python language. The course consists of four content units and one completion (summary) test. The units include (i) introduction to Python and computer programming, (ii) data types variables, basic I/O, operations and basic operators, (iii) boolean values, conditional loops, lists, logical and bitwise operators, and (iv) functions, tuples, dictionaries, data processing and exceptions.

\emph{Python Essentials - Part 2 (Intermediate)} (\textbf{PE2})\footnote{OpenEDG: Python Essentials - Part 2 (Intermediate). Available at: \url{https://edube.org/study/pe2} [Accessed 2023-01-15]} is focused on more advanced aspects of Python programming, including modules, packages, exceptions, file processing, object-oriented programming. Similarly to PE1, the course is organized into four content units and one completion (summary) test. The course units are (i) modules, packages, and pip, (ii) strings, string and list methods, and exceptions, (iii) object-oriented programming, and (iv) miscellaneous. 

Finally, \emph{Practical Programming with Python}\footnote{Sail(): Social and Interactive Learning Platform. Available at: \url{https://sailplatform.org/courses}. [Accessed 2023-03-03]} (\textbf{PPP}) emphasizes hands-on experience with fundamental Python constructs and exposure to software development tools, practices, and real-world applications. The course consists of eight units which include (i) Python basics and introduction to functions, (ii) control flow, strings, input and output, (iii) Python data structures, (iv) object-oriented programming, (v)~software development, (vi) data manipulation, (vii) web scraping and office document processing, and (viii) data analysis.

In PE1 and PE2, formative assessments are called quizzes while summative assessments are called tests. The tests determine if learners pass the courses whereas quizzes are meant as practice. The MCQs often include small snippets of code for learners to reason about. From the two courses, we collected 297 questions (179 have code snippets). PPP uses MCQ-style inline activities as formative assessment and tests as summative assessment. From this course, we collected 233 MCQs (144 with code snippets). Table~\ref{tab:dataset} has additional details.

\begin{table}
  \caption{Descriptive statistics of the created dataset. Each row provides information about the MCQs each of the courses employ. Each column reports on the distribution of the code content of each MCQ set in each course.}
  \label{tab:dataset}
  \setlength{\tabcolsep}{5pt}
  \begin{tabular}{l|r|rrr}
  \hline
    Course        & Units    & MCQ     & MCQ     & Course\\
                  & (topics) & (plain) & (+code) & Overall \\
  \hline
    PE1           & 4        & 53          & 96          &\bf 149  \\
    PE2           & 4        & 65          & 83          &\bf 148  \\
    PPP           & 8        & 89          & 144         &\bf 233  \\
  \hline
    Type Overall  & 16       & \bf 207     & \bf 323      &\bf 530  \\
  \hline
  \end{tabular}
\end{table}

We used simple pattern matching combined with manual curation as the second step to organize the MCQs into several categories. The first distinction was made between MCQs \emph{with code} and MCQs with \emph{no code}. For an MCQ, to be considered as \emph{with code} one of the following two had to be true:

\begin{itemize}
    \item Within the body of the question there had to be at least one line fully dedicated to computer code.
    \item The choices were computer code expressions.
\end{itemize}

\noindent Inline mentions of names of functions or variables were not considered as sufficient for an MCQ to be considered \emph{with code}.

\begin{figure}[t]
\includegraphics[width=.5\textwidth]{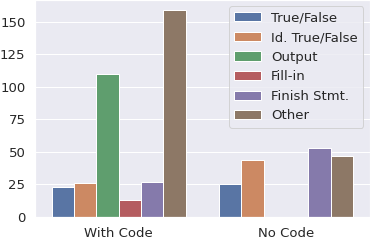}
\caption{Distribution of MCQs into categories. Note that the MCQs asking about the output of a code snippet as well as MCQs focused on filling-in the blanks in a snippet are not present in the MCQs with no code. This is to be expected given the nature of those questions. The MCQs with code are quite dominated by questions that ask about the output of a code snippet as well as with questions of other type. Otherwise, the distribution is relatively uniform.}
\label{fig:mcq_categories_dist}
\end{figure}

The second distinction was made along the following lines, focusing on the overall syntax of what the question writer asks the student to do:

\begin{itemize}
    \item \textbf{True/False}\\ The learner is asked to assess the truthfulness of a single statement. For example:
    \begin{quote}
        Developers that write code individually are not expected to apply code standards.\\
        A. True\\
        \emph{B. False}
    \end{quote}
    \begin{quote}
        Evaluate the following expression and determine whether it is True or False.
        \begin{verbatim}2 + 2 != 2 * 2\end{verbatim}
        A. True\\
        \emph{B. False}
    \end{quote}
    \item \textbf{Identify True/False Statement}\\
    The learner is asked to pick one or more answer choices that are either true or false. Note that this is different from the True/False questions (previous category). For example:
    \begin{quote}
        Which of the following statements is false?\\
        A. The pandas module provides some CSV-related methods.\\
        B. Python has a built-in XML package with several modules for XML parsing.\\
        \emph{C. JSON data format has syntax to represent all Python data structure types.}\\
        D. Python has a built-in \verb|csv| module containing methods for reading and writing into CSV files.
    \end{quote}
    \begin{quote}
        Take a look at the snippet and choose one of the following statements which is true:
        \begin{verbatim}
 nums = []
 vals = nums[:]
 vals.append(1)\end{verbatim}
         A. \verb|nums| is longer than `vals`\\
         \emph{B.} \verb|vals| \emph{is longer than} \verb|nums|\\
         C. \verb|nums| and \verb|vals| are of the same length
    \end{quote}
    \item \textbf{Finish Statement.}\\
    The learner is asked to complete a statement. For example:
    \begin{quote}
        The `**` operator:\\
        A. performs duplicated multiplication\\
        B. does not exist\\
        \emph{C. performs exponentiation}
    \end{quote}
    \begin{quote}
        Right-sided binding means that the following expression:
       \begin{verbatim}1 ** 2 ** 3\end{verbatim}
        will be evaluated:\\
        \emph{A. from right to left}\\
        B. in random order\\
        C. from left to right
    \end{quote}
    \item \textbf{Output}\\
    The learner is asked to identify the choice that corresponds to the output of a given snippet of code. This category is applicable only to questions \emph{with code}. For example:
    \begin{quote}
        What is the output of the following snippet if the user enters two lines containing \verb|2| \emph{and} \verb|4| \emph{respectively?}
\begin{verbatim}
x = int(input())
y = int(input())
print(x + y)\end{verbatim}
    A. 2\\
    B. 24\\
    \emph{C. 6}
    \end{quote}
    \begin{quote}
        What is the output of the following snippet?
\begin{verbatim}
my_list_1 = [1, 2, 3]
my_list_2 = []
for v in my_list_1:
my_list_2.insert(0, v)
print(my_list_2)\end{verbatim}
        A. \verb|[1, 2, 3]|\\
        B. \verb|[1, 1, 1]|\\
        C. \verb|[3, 3, 3]|\\
        \emph{D.} \verb|[3, 2, 1]|
    \end{quote}
\item \textbf{Fill-in Blanks}\\
The learner is asked to fill in a code snippet by selecting the appropriate choice as an answer. This category is applicable only to questions \emph{with code}. For example:
    \begin{quote}
    Fill in the blank of the \verb|is_negative| function definition shown below, so that the function returns \verb|True| when the argument provided to \verb|num| is a negative number and returns \verb|False| otherwise.
\begin{verbatim}
def is_negative(num):
    return _________________
\end{verbatim}
    A. \verb|not (num > 0)|\\
    B. \verb|num > 0|\\
    C. \verb|num <= 0|\\
    \emph{D.} \verb|num < 0|
    \end{quote}
    \begin{quote}
        The following code snippet should open the \verb|myfile| file and assign the lines to the \verb|all_lines| variable. Which of the options below should be used to fill in the blanks?
 
\begin{verbatim}
with __________________________
    all_lines = file.readlines()\end{verbatim}
    \emph{A.} {\ttfamily \emph{open("myfile",'r') as file:}}\\
    B. \verb|"myfile" in open as file:|\\
    C. \verb|with open "myfile" as file:|
    \end{quote}
    \item \textbf{Other}\\
    Any MCQ that does not fall into any of the above categories. For example:
    \begin{quote}
    How many times will the code snippet below print `X`.?
 
\begin{verbatim}
for i in range(1, 7):
    for j in range(2, 6):
    print('X')\end{verbatim}
\emph{A. 24}\\
B. 28\\
C. 35
    \end{quote}
Notice that the above example is closely related to the questions asking for the \emph{output} of the snippet. However, there is a subtle difference since this questions does not ask what is the output directly.
    \begin{quote}
        Given the piece of code presented in the code snippet below, what is the value of \verb|palindromes[1]|?
\begin{verbatim}
palindromes = ['pop', 'noon', 'madam']
\end{verbatim}
    A. \verb|'pop'|\\
    \emph{B.} \verb|'noon'|\\
    C. \verb|'p'|\\
    D. \verb|'madam'|\\
    E. \verb|'o'|
    \end{quote}
\end{itemize}

\noindent Figure \ref{fig:mcq_categories_dist} shows the distribution of the MCQs into the individual categories. The MCQs asking about the output of a code snippet as well as MCQs focused on filling-in the blanks in a snippet are not present in the MCQs with \emph{no code}. This is to be expected given the nature of those questions. The MCQs with code are quite dominated by questions that ask about the \emph{output} of a code snippet as well as with questions of \emph{other} type. Otherwise, the distribution is relatively uniform. The \emph{fill-in} questions are rare. The distribution of the \emph{no code} questions is close to uniform.

\section{\uppercase{Models}}
\label{sec:models}
The original GPT model \cite{radford2018improving} is a 12-layer decoder-only transformer \cite{vaswani2017attention} with masked self-attention heads. Its core capability is fine-tuning on a downstream task. The GPT-2 model~\cite{radford2019language} largely follows the details of the original GPT model with a few modifications, such as layer normalization moved to the input of each sub-block, additional layer-normalization after the first self-attention block, and a modified initialization. Compared to the original model it displays remarkable multi-task learning capabilities~\cite{radford2019language}. The next generation of GPT models~\cite{brown2020language} uses almost the same architecture as GPT-2. The only difference is that it uses alternating dense and locally banded sparse attention patterns in the layers of the transformer. The main focus of Brown et al. was to study the dependence of performance and model size where eight differently sized models were trained (from 125 million to 175 billion parameters). The largest of the models is commonly referred to as GPT-3. The interesting property of these models is that they appear to be very strong zero- and few-shot learners. This ability appears to improve with the increasing size of the model~\cite{brown2020language}.

We are primarily interested in the performance of \verb|text-davinci-003|, one of the most advanced GPT models offered by OpenAI. The \verb|text-davinci-003| model builds on top of previous \verb|text-davinci-002|, which in turn is based on \verb|code-davinci-002| (focused on code-completion tasks). To gauge the rate of improvement over the several recent years, we compare the performance of \verb|text-davinci-003| to \verb|text-davinci-002| as well as to the previous generation's InstructGPT model (\verb|text-davinci-001|).\footnote{OpenAI: Model index for researchers. Available at: \url{https://beta.openai.com/docs/model-index-for-researchers/instructgpt-models} [Accessed 2023-01-15]} Recently, OpenAI has also released \verb|gpt-3.5-turbo| which reportedly matches the performance of the \verb|text-davinci-003| for tenth of the cost. 

We set the \verb|temperature| to 0.0, which corresponds to no randomness. The higher the \verb|temperature| the more creative the output but it can also be less factual. We set \verb|max_tokens| to 500 (a token roughly corresponds to a word). This parameter controls the maximum length of the output. We set \verb|top_p| to 1, as is recommended  when \verb|temperature| is set to 0.0. This parameter is related to \verb|temperature| and also influences creativeness of the output.  We set \verb|frequency_penalty| to 0, which allows repetition by ensuring no penalty is applied to repetitions. Finally, we set \verb|presence_penalty| to 0, ensuring no penalty is applied to tokens appearing multiple times in the output. 


\section{\uppercase{Experimental Design}}
\label{sec:experimental_design}
To test the performance of the three \verb|text-davinci-*| models, we submit MCQs one by one using the \verb|openai| Python library\footnote{GitHub: OpenAI Python Library. Available at: \url{https://github.com/openai/openai-python} [Accessed 2023-01-16]} which is a wrapper for the OpenAI's REST API. We embed each question in the prompt template shown in Figure \ref{fig:mcq-prompt-template}. The text of the prompt's preamble is inspired by OpenAI's QA example.\footnote{OpenAI: Q\&A. Available at: \url{https://platform.openai.com/examples/default-qa} [Accessed 2023-03-04]} The 
\{\{question\}\} token is replaced with the question text. The \{\{choices\}\} token is replaced with the candidate answers where each one is placed on a single line preceded by a capital letter. Each model returns one or more of the choices as the prompt completion, which is then compared to the reference answer. For PE1 and PE2, we let partially correct answers be incorrect, following the course creators' assessment guidelines. In PPP, there is always exactly one correct answer.


As the baseline, we use a simple model that selects the answer with the highest Jaccard similarity to the question. In case of a tie the longest answer is selected. Jaccard similarity is one of the simplest measures of text similarity. Hence, it is an ideal candidate for a baseline as it allows to detect what ratios of the questions within their respective categories could be solved employing this simple, yet sensible, heuristic. Such MCQs likely pose very little challenge for GPT models.


We report the proportions of the correct answers~(i.e., the accuracy) for each model per MCQ category. We specifically focus on the differences in performance of the \verb|text-davinci-003| model on MCQs that contain code snippets (\emph{with code}) compared to MCQs that do not (\emph{no code}). We are also interested in the difference between the performance on completion-based MCQs (\emph{Finish Statement} and \emph{Fill-in Blanks}) compared to the rest. This is because these question types are not too far off from the pre-training objective and, hence, the expectation is that the models' performance should be higher on these types. To test statistical significance we use a simple two-independent proportions test which is a statistical hypothesis test used to determine whether two proportions are different from each other. 

\begin{figure}
\footnotesize
\begin{Verbatim}[frame=single,commandchars=\\\{\}]
I am a highly intelligent bot that can easily
handle answering multiple-choice questions on 
introductory Python topics. Given a question 
and choices I can always pick the right ones.

Question: \textcolor{orange}{\string{\string{question\string}\string}}

Choices:
\textcolor{orange}{\string{\string{choices\string}\string}}

The correct answer:
\end{Verbatim}
\begin{textblock*}{3.4cm}(5.7cm,-2.75cm)
\circledorange{1}
\end{textblock*}
\begin{textblock*}{3.4cm}(3.7cm,-2.45cm)
\circledorange{2}
\end{textblock*}
\begin{textblock*}{3.4cm}(1.95cm,-1.4cm)
\circledorange{3}
\end{textblock*}
\caption{MCQ Prompt Template. The text of the preamble (1) is inspired by OpenAI's QA example. The 
\{\{question\}\} token~(2) is replaced with the question text. The \{\{choices\}\} token~(3) is replaced with the candidate answers where each one is placed on a single line preceded by a capital letter.}
\label{fig:mcq-prompt-template}
\end{figure}

\section{\uppercase{Results}}
\label{sec:results}

\begin{table*}[t!]
  \caption{Results of the experiments. The Jaccard column reports the performance of the baseline. The text-davinci-001, text-davinci-002, and text-davinci-003 columns report the performance of the different GPT3 models. Results of the No Code and With Code sections are summarized in the Total rows. The Overall row at the bottom reports the average performance of the models across all the types of MCQs.}
  \label{tab:results}
  \centering
  \begin{tabular}{lrrrr}
  \hline
    Question Type & Jaccard & text-davinci-001  & text-davinci-002     & text-davinci-003 \\
  \hline
    \multicolumn{5}{c}{\bf No Code}      \\
  \hline
    True/False     & 11/25 & 13/25   & 19/25   & 20/25   \\
                   & (44.0\%) & (52.0\%)& (76.0\%)& (80.0\%)\\
    Identify True/False Statement & 8/44   & 12/44   & 22/44   & 27/44   \\
                   & (18.2\%)& (27.3\%)& (50.0\%)& (61.4\%)\\
    Finish Statement   & 12/53   & 40/53   & 46/53   & 48/53   \\
                   & (22.6\%)& (75.5\%)& (86.8\%)& (90.6\%)\\
    Other          & 9/47   & 27/50   & 43/50   & 39/50   \\
                   & (19.1\%)& (53.2\%)& (86.0\%)& (74.0\%)\\
    \bf Total      &\bf 40/172  &\bf 92/172  &\bf 130/172 &\bf 134/172   \\
                   &\bf (23.2\%)&\bf (53.5\%)&\bf (75.6\%)&\bf (77.9\%)\\
  \hline
    \multicolumn{5}{c}{\bf With Code}      \\
  \hline
    True/False     & 9/23   & 12/23   & 10/23   & 10/23   \\
                   & (39.1\%)& (52.2\%)& (43.5\%)& (43.5\%)\\
    Identify True/False Statement & 4/26   & 10/26   & 15/26   & 11/26   \\
                   & (15.4\%)& (38.5\%)& (57.7\%)& (42.3\%)\\
    Output         & 22/110  & 28/110  & 58/110   & 53/110  \\
                   & (20.0\%)& (25.4\%)& (52.7\%)& (48.2\%)\\
    Fill-in        & 2/13    & 5/13    & 10/13   & 11/13   \\
                   & (15.4\%)& (38.5\%)& (76.9\%)& (84.6\%)\\
    Finish Statement   & 8/27   & 10/27   & 22/27   & 22/27   \\
                   & (29.6\%)& (37.0\%)& (81.5\%)& (81.5\%)\\
    Other          & 39/159  & 42/159  & 97/159  & 106/159\\
                   & (24.5\%)& (26.4\%)& (61.1\%)& (66.7\%)\\
    \bf Total      &\bf 84/358 &\bf 107/358 &\bf 212/358 &\bf 213/358   \\
                   &\bf (23.4\%)&\bf (29.9\%)&\bf (59.2\%)&\bf (59.5\%)\\
  \hline
    Overall        & 124/530 & 199/530 & 342/530 & 347/530   \\
                   & (23.4\%)& (37.5\%)& (64.5\%)& (65.5\%)\\
  \hline
  \end{tabular}
\end{table*}

Table \ref{tab:results} reports the results of our experiments. Firstly, as one would expect all three GPT models clearly outperform the simple Jaccard similarity baseline. The \verb|text-davinci-003| model appears to perform the best (65.5\% overall) with a small margin over the \verb|text-davinci-002| (64.5\% overall). The performance of the \verb|text-davinci-001| appears to be much lower compared to the other two models. This is to be expected. While the \verb|text-davinci-002| is a direct predecessor of the \verb|text-davinci-003| (hence, the small difference) the \verb|text-davinci-001| is quite removed from the two. The major breakthrough in OpenAI GPT-3's capabilities in handling computer code was Codex (\verb|code-davinci-002|) \cite{https://doi.org/10.48550/arxiv.2107.03374} which is the direct predecessor of \verb|text-davinci-002|.\footnote{OpenAI: Model index for researchers. Available at: \url{https://beta.openai.com/docs/model-index-for-researchers/instructgpt-models} [Accessed 2023-01-15]}

There appears to be a clear difference between the performance of the most capable \verb|text-davinci-003| on the MCQs that contain code snippets (59.5\% overall) compared to those that do not (77.9\% overall). This difference is statistically significant ($p<0.0001$). This is to be expected as the combination of code and natural language likely constitutes (on average) more complex input than natural language alone. Additionally, it is quite possible that in our particular context the questions with code are (on average) more difficult than questions with no code.

There also appears to be clear difference between the performance of \verb|text-davinci-003| on the completion-oriented MCQs (87.1\%) and the rest (60.1\%). This difference is statistically significant ($p<0.0001$). Since GPT models are primarily focused on prompt completion, be it text or computer code, this finding is also as expected.



\section{\uppercase{Discussion}}
\label{sec:discussion}
Our experimental results suggest that there, indeed, is a difference between how successfully the GPT models handle questions that contain only natural language and those that also contain snippets of computer code (RQ1). Tentatively, we can conclude that inclusion of a code snippet within an MCQ makes the question more challenging for GPT models to handle. This conclusion is supported by universally lower performance on MCQs with code across all the subtypes, i.e., \emph{True/False}, \emph{Identify True/False Statement}, \emph{Finish Statement}, and \emph{Other}. The root cause for this discrepancy is likely one or more of the following: (i)~GPT models are somewhat more limited with respect to handling computer programs compared to natural language; (ii) GPT models struggle with the combination of different types of expressions (i.e., natural language and code); and/or (iii) the questions with code snippets are inherently more difficult.

While the greater difficulty of the questions with code might certainly be a factor it appears that the GPT models sometimes struggle to answer questions with code that one might judge as simple. For example, consider the following MCQ:

\begin{quote}
    The following statement:
\begin{verbatim}
assert var == 0\end{verbatim}
A. is erroneous\\
\emph{B. will stop the program when var != 0}\\
C. has no effect\\
D. will stop the program when var == 0
\end{quote}

\noindent The answer of \verb|text-davinci-003| to this question was ``D. will stop the program when var == 0''. Hence, it appears there are certain limitations in the capabilities of the GPT models to answer questions about code. This is somewhat surprising if one considers the well documented capabilities of the models when it comes to generation or explanation of computer programs.

The results also show that certain types of MCQs are more challenging than others for the GPT models (RQ2). The questions that involve generation of natural language and/or code appear to be handled with much more success than other types of questions. This is to be expected as GPT models are primarily focused on prompt completion. On the other hand, it leads to somewhat paradoxical situations such as the one illustrated in the motivating example (Section \ref{sec:motivation}). The models are capable of generating code based on a natural language description, as well as generating natural language explaining execution of the code line-by-line. Yet, somehow these capabilities do not seem to extend to the realm of answering pointed specific questions about the code (often quite simple ones).

We hypothesize that the above described paradox might be related to the phenomenon described by \cite{Detienne2002}. They point out that program code serves two simultaneous purposes: it is both a narrative description of a programmer's intent, and an artifact that controls computer execution. Accordingly, human programmers maintain, and synchronize, at least two kinds of mental models of code, a \emph{functional} model that captures the purpose the program is supposed to play in the world, and a \emph{structural} model that allows mental simulation of data and control flow.

Since GPT models are trained on large corpora that include texts in natural language as well as program code with comments and documentation, they may acquire robust representations of the functional relationship between code and the intent it expresses.
The training corpora likely do not contain code with outputs or trace logs of its execution. Thus, models may lack the required data to build a representation of a structural model of code's function. This is not to say that including the mentioned resources into the training corpora would necessarily result in the acquisition of such a model. This is because an effective use of the model may require the ability to simulate execution of the code, or its part. The current large language models, including GPT, do not have this capability. Note that there is an active area of research in augmenting large language models with reasoning
skills and providing them with the ability to use tools, such as the Python interpreter \cite{mialon2023augmented}.

Arguably, building up these connected mental models of code's purpose and operation should be a key part of what CS education teaches. The particular limitations of GPT models provide a useful lens into what kind of mental model we are evaluating in typical higher education programming assessments. It may be that \emph{True/False} and \emph{Identify True/False Statements} MCQs more likely require mental simulation of the code execution. An experiment to validate our hypothesis might be to classify MCQs according to their focus on (a) predicting actual behavior, or (b) inferring intent, and measure if and how the GPT models' performance correlates with this classification. 

There are ongoing debates as to the changes the emergence of GPT-based tools such as ChatGPT or GitHub Copilot will inflict on the software development profession as well as programming education. Firstly, it appears inevitable that the tools will become an integral and accepted part in the software development process. Therefore, future programmers will likely need to write less code. On the other hand, they will need to be able to validate the auto-generated code, spot deficiencies, and correct them efficiently. Hence, programming education might need to de-prioritize teaching learners how to write code and start emphasizing skills such as requirements
formulation, debugging, trade-off analysis, and critical thinking.

Finally, the GPT-based tools present numerous opportunities to improve current instructional and assessment practices in programming classes. Our experiments suggest that GPT models are capable of explaining code in plain and easily understandable terms. Similarly, they are capable of generating and completing program code. A judicious use of these capabilities might result in numerous novel tools and instructional approaches for novices and advanced learners alike. However, there are also potential threats. An improper or misinformed use of the tools may result in an academic integrity violation (AIV) incident (i.e., cheating). Similarly, over-reliance on GPT-based tools may rather hinder than improve the learning process.

\section{\uppercase{Conclusions and Future Work}}
\label{sec:conclusions}
We evaluated \verb|text-davinci-*| GPT models on a sizeable set of 530 MCQs, many of which contained code snippets, from three Python programming courses. The overall accuracy of the most capable \verb|text-davinci-003| model was measured at 65.5\%~(compared to the 23.4\% Jaccard similarity baseline). While such performance is impressive there appear to be some noticeable limitations. First of all, it appears that the MCQs containing code snippets were somewhat more challenging (59.5\%) for the model than those with no code (77.9\%). In addition, MCQs that ask to complete a sentence or fill-in a blank appear to be handled much more successfully (87.1\%) compared to other types of questions~(60.1\%). Therefore, GPT models' capabilities seem limited when it comes to handling MCQs about computer code requiring reasoning beyond mere completion (56.6\%).

While our study of GPT models' performance on diverse types of MCQs yielded numerous valuable insights, it is subject to countless limitations and leaves much room for improvement. Hence, we suggest several directions for future work: (i)~further analyze the effects of prompt-tuning (ii) and/or iterative prompt-construction; (iii) examine the performance of GPT models on other domains, e.g., competitive mathematics; (iv) develop a systematic framework to comprehensively assess the capabilities and limitations of GPT models; and (v)~study possibilities of effective integration of GPT-based tools, e.g., ChatGPT or Copilot, into programming education.




\bibliographystyle{apalike}
{\small
\bibliography{refs}}



\end{document}